%% file: example_paper.tex
\documentclass{article}

\usepackage{microtype}
\usepackage{graphicx}
\usepackage{subcaption}
\usepackage{booktabs} 
\usepackage{multirow}
\usepackage{array}
\usepackage{tabularx}
\usepackage{caption}
\usepackage{amsmath}
\usepackage{amssymb}
\usepackage{algorithm}
\usepackage{algorithmic}

\usepackage{hyperref}

\usepackage[preprint]{icml2026}

\usepackage{amsmath}
\usepackage{amssymb}
\usepackage{mathtools}
\usepackage{amsthm}

\usepackage[capitalize,noabbrev]{cleveref}

\theoremstyle{plain}

\theoremstyle{definition}

\theoremstyle{remark}

\usepackage[textsize=tiny]{todonotes}

\begin{document}

\twocolumn[
  \icmltitle{Active-SAOOD: Active Sparsely Annotated Oriented Object Detection in Remote Sensing Images}



  \icmlsetsymbol{equal}{*}
  \begin{icmlauthorlist}
    \icmlauthor{Yu Lin}{xmu}
    \icmlauthor{Jianghang Lin}{xmu}
    \icmlauthor{Kai Ye}{xmu}
    \icmlauthor{Shengchuan Zhang}{xmu}
    \icmlauthor{Liujuan Cao $^\dagger$}{xmu}
  \end{icmlauthorlist}

  \icmlaffiliation{xmu}{Key Laboratory of Multimedia Trusted Perception and Efficient Computing, Ministry of Education of China, Xiamen University, 361005, P.R. China.}

  \icmlcorrespondingauthor{Liujuan Cao}{caoliujuan@xmu.edu.cn}


  \vskip 0.3in
]



\printAffiliationsAndNotice{}  

\input{sec/0_abstract}
\input{sec/1_intro}
\input{sec/2_related_work}
\input{sec/3_method}
\input{sec/4_experiments}
\input{sec/5_conclusion}

\bibliography{example_paper}
\bibliographystyle{icml2026}

\newpage
\appendix
\renewcommand{\thesection}{\arabic{section}}
\renewcommand{\thesubsection}{\thesection.\arabic{subsection}}
\clearpage


\twocolumn[
  \icmltitle{Active-SAOOD: Active Sparsely Annotated Oriented Object Detection in Remote Sensing Images (Supplementary Material)}
]
\input{sec/X_suppl}

\end{document}

%% file: sec/0_abstract.tex
\begin{abstract}
Reducing the annotation cost of oriented object detection in remote sensing remains a major challenge. Recently, sparse annotation has gained attention for effectively reducing annotation redundancy in densely remote sensing scenes. However, (1) the sparse data reliance on class-dependent sampling, and (2) the lack of in-depth investigation into the characteristics of sparse samples hinders its further development. This paper proposes an active learning-based sparsely annotated oriented object detection (SAOOD) method, termed Active-SAOOD. Based on a model state observation module, Active-SAOOD actively selects the most valuable sparse samples at the instance level that are best suited to the current model state, by jointly considering orientation, classification, and localization uncertainty, as well as inter- and intra-class diversity. This design enables SAOOD to operate stably under completely randomly initialized sparse annotations and extends its applicability to broader real-world. Experiments on multiple datasets demonstrate that Active-SAOOD significantly improves both performance and stability of existing SAOOD methods under various random sparse annotation. In particular, with only 1\% annotated ratios, it achieves a 9\% performance gain over the baseline, further enhancing the practical value of SAOOD in remote sensing. The code will be public.
\end{abstract}

%% file: sec/1_intro.tex
\section{Introduction}
\label{sec1}
Oriented object detection (OOD) in remote sensing images is a fundamental task with wide applications such as port management, traffic planning, and agricultural monitoring \cite{S2A_Net, R3Det}, but its application is hindered by the high cost of manual annotation. To reduce the labeling cost, recent works have explored weakly supervised \cite{H2RBox}, semi-supervised \cite{SOOD}, and sparsely annotated \cite{RSST} approaches. Sparsely annotated oriented object detection (SAOOD) has attracted increasing attention for well reducing annotation redundancy in dense remote sensing scenes.
\input{Figure/img1}

Nevertheless, most existing SAOOD methods rely on strong category-dependent priors. Some construct sparse annotations by sampling a fixed proportion of instances per category \cite{PECL}, which requires prior knowledge of the global data distribution, while others select samples based on per-image class statistics \cite{DML}, demanding careful image-level labeling to avoid missing categories. These assumptions limit practical scalability. Under fully random sparse annotation, current SAOOD methods suffer significant performance degradation, as many low-value samples consume annotation budgets while providing limited training benefit, resulting in low annotation efficiency.

Although active learning has been extensively studied in image classification \cite{li2024survey} and semi-supervised object detection \cite{Active_teacher}, existing approaches mainly explore image-level annotation value by measuring image difficulty or diversity to estimate the information richness of samples. However, SAOOD requires active learning for sparsely annotated instances rather than entire images, and instance-level active learning can select high-value instances under the same annotation budget, thereby improving model performance. To the best of our knowledge, instance-level active learning tailored for SAOOD has not yet been explored. It has three key differences compared with image-level active learning: (1) Compared with image-level active learning, instance-level active learning must consider more fine-grained evaluation dimensions, such as instance-level localization uncertainty and intra-class diversity. (2) Oriented object detection introduces an additional angle prediction, whose impact on instance annotation value remains underexplored. (3) Image-level active learning can directly select the most difficult images, since each image contains multiple instances, ensuring diverse instance difficulties, whereas instance-level active learning cannot simply select the hardest instances, as heavily sampling difficult instances may harm training stability in early stages, which is widely demonstrated in curriculum learning \cite{curriculum_learning}.

These observations indicate that existing image-level active learning strategies cannot be directly applied to SAOOD task. To address these challenges, we propose Active-SAOOD, a fine-grained instance-level active learning framework for sparsely annotated oriented object detection. First, to overcome the granularity mismatch, we start from the intrinsic nature of the SAOOD and decompose the task into classification, localization, and orientation, defining corresponding instance-level uncertainties to select samples that enhance model capability along each dimension. We further incorporate inter- and intra-class diversity to handle the long-tailed distribution and large intra-class variations in remote sensing images, thereby improving detection of rare categories and intra-class robustness. Second, to address the previously unexplored orientation dimension, we introduce orientation uncertainty to identify instances that are particularly challenging for angle prediction. Third, to mitigate the instability caused by naively selecting the hardest instances, we design a model state observation–based selection strategy that dynamically adjusts the weighting of different evaluation criteria according to the current model capability, gradually increasing the difficulty of selected samples as training progresses. Together, these components constitute a comprehensive instance-level active learning evaluation framework, enabling Active-SAOOD to substantially improve both detection performance and training stability under extremely sparse annotations, thereby extending SAOOD to more practical scenarios. Our main contributions are summarized as follows:

\begin{itemize}
\item[1)] To mitigate the performance degradation of SAOOD caused by randomly sparse annotations, we propose Active-SAOOD, an active learning–based sparsely annotated oriented object detection framework. To address the limitation that conventional image-level active learning is not well suited for sparse annotation settings, Active-SAOOD introduces a novel instance-level active learning paradigm. It actively selects the most suitable high-value samples that best match the model state for annotation, thereby improving both performance and training stability.
\item[2)] To better adapt to the different capability requirements of the model at different training stages, we introduce a model state observation–based selection strategy that assesses the capability of model and guides active selection of high-value instances tailored to the current model state, enabling effective integration of active learning with sparse annotation and extending SAOOD to more practical scenarios.
\item[3)] Extensive experiments show that Active-SAOOD greatly improves SAOOD performance under random sparse initialization, achieving a 9\% gain over the baseline at 1\% annotation and boosting training stability.
\end{itemize}

%% file: Figure/img1.tex
\begin{figure}[t]
    \centering
    \includegraphics[width=\linewidth]{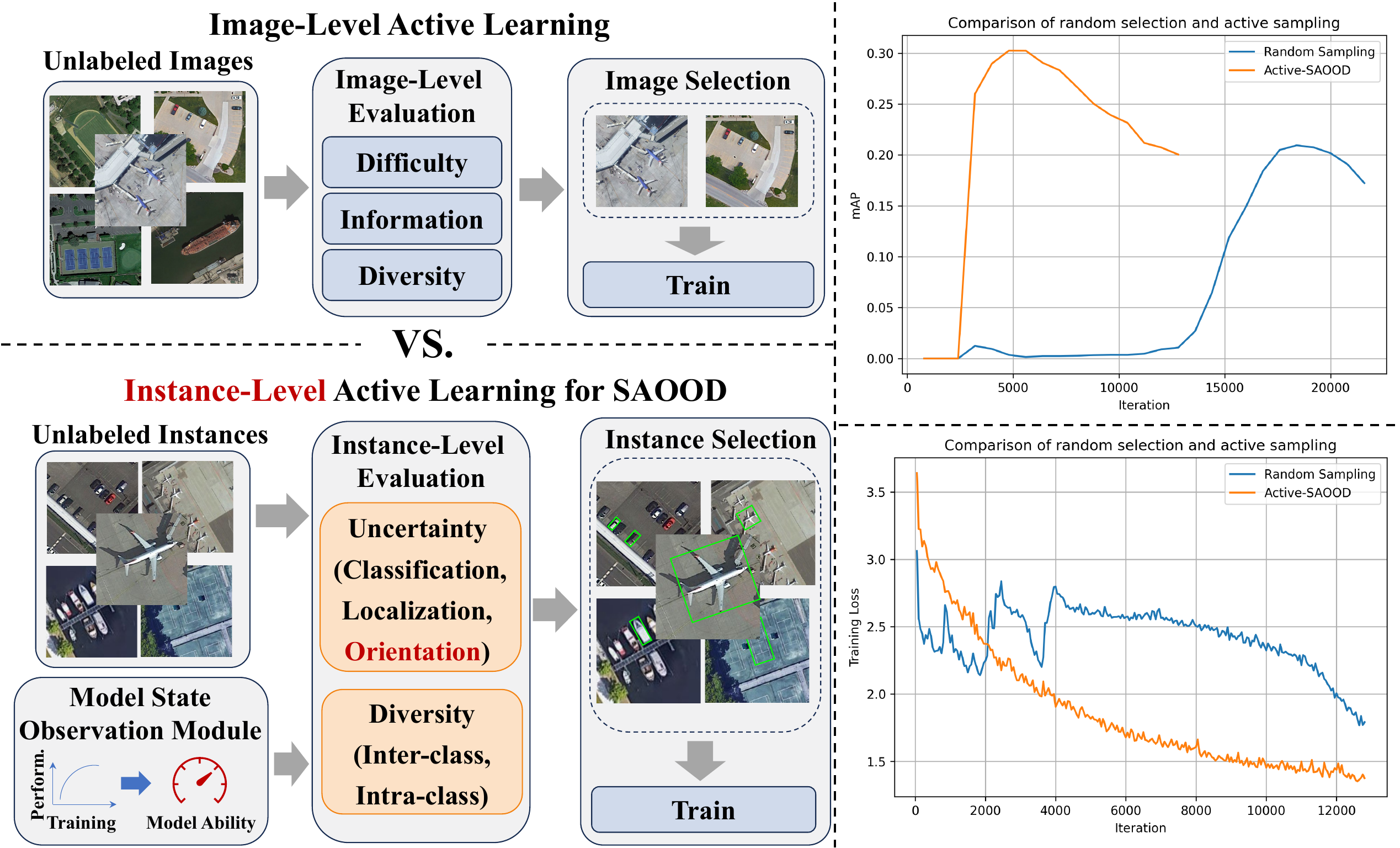}
    \caption{Comparison of different sampling methods. Image-level active learning selects high-value images based on image-level evaluation criteria. Instance-level active learning considers richer instance-level evaluation dimensions and selects instances according to the model state. Compared to random sparse annotation of many low-value samples, our Active-SAOOD achieves higher performance via the instance-level active learning.}
    \label{Fig1}
    \vspace{-5pt}
\end{figure}

%% file: sec/2_related_work.tex
\section{Related Work}
\label{sec2}
\textbf{Oriented Object Detection.} Oriented object detection (OOD) predicts rotated boxes (RBoxes) rather than horizontal ones, explicitly modeling object orientation, and is widely used in scene text and remote sensing \cite{he2017deep, Oriented_RCNN}. Early OOD methods \cite{RRPN} extended horizontal detectors, while later works tailored for remote sensing improved feature representation (R$^3$Det \cite{R3Det}), anchor quality (S$^2$A-Net \cite{S2A_Net}), and proposal efficiency (Oriented R-CNN \cite{Oriented_RCNN}). Other approaches address boundary discontinuity \cite{KFIoU, PSC}, learn rotation-equivariant features \cite{ReDet}, or enlarge receptive fields \cite{LSKNet}. Despite these advances, OOD remains annotation-intensive, motivating research on low-cost annotation methods.

\textbf{Sparsely Annotated Oriented Object Detection.} Recently, sparsely annotated oriented object detection (SAOOD) has gained attention for reducing annotation redundancy in dense remote sensing scenes. PECL \cite{PECL} used conformal learning for high-quality pseudo-labels, DML \cite{DML} applied dynamic multi-view learning, S$^2$Teacher \cite{S2Teacher} leveraged angle-consistency and progressive mining, and RSST \cite{RSST} utilized large language models for pseudo-label assignment. However, these methods rely on strong category-dependent priors, making random initialization unstable, and the value of sparse samples is largely unexplored. To address this, we propose Active-SAOOD, which actively selects high-value samples for sparse annotation, ensuring stable performance and extending SAOOD to more general scenarios.

\textbf{Active Learning.} Active learning reduces annotation costs by selecting the most informative samples. While it has been widely studied in image classification \cite{wang2016cost} and extended to semi-supervised object detection, Active Teacher \cite{Active_teacher} selects valuable images, MI-AOD \cite{MI_AOD} estimates uncertainty via multiple instance learning, ALWOD \cite{ALWOD} uses teacher–student disagreement, and PPAL \cite{PPAL} provides a plug-and-play module. These methods focus on image-level value. Sparse annotation, however, emphasizes instance-level value, and existing approaches ignore orientation uncertainty, which is critical in OOD. We address this by introducing multi-dimensional instance-level annotation value, including orientation, enabling active learning to enhance SAOOD under random initialization.

%% file: sec/3_method.tex
\section{Active-SAOOD}
Given a small sparsely annotated instances set $\mathcal{D}_I=\{\mathcal{X}_I,\mathcal{Y}_I\}$ and a large unlabeled instance set $\mathcal{D}_U=\{\mathcal{X}_U\}$, where $\mathcal{X}$ denotes instances and $\mathcal{Y}$ denotes labels, Active-SAOOD aims to select the most valuable instances from $\mathcal{D}_U$ for annotation and merge them with $\mathcal{D}_I$ to form an actively annotated set $\mathcal{D}_A$, which is used to train the oriented object detector for improving performance.
\input{Figure/img2}

As shown in Fig. \ref{Fig2}, Active-SAOOD starts from a small set of randomly and sparsely annotated instances to initialize a teacher–student SAOOD framework. The initialized teacher model evaluates the annotation value of samples based on instance-level classification, localization, and orientation uncertainties, as well as inter- and intra-class diversity. Guided by the model state observation strategy, it selects the samples that are most suitable for the current model state. The selected samples are then manually annotated and merged with the initial training set to form the active learning training set, which is subsequently used to iteratively train the student model.

\subsection{Instance-Level Active Sampling for SAOOD}
Existing active learning methods have mainly focused on image classification and semi-supervised object detection, and cannot be directly applied to SAOOD for two reasons. First, they target image-level annotation value, whereas SAOOD emphasizes instance-level value, creating a granularity mismatch. Second, the angular dimension in oriented object detection and its impact on annotation value remain unexplored. To address this, we decompose SAOOD into classification, localization, and orientation tasks, and define instance-level uncertainties for each dimension to select samples that improve the corresponding model capabilities. To handle the long-tailed distribution and large intra-class variations in remote sensing imagery, we further introduce inter- and intra-class diversity, encouraging selection of samples that enhance detection of rare categories and intra-class robustness. Together, these criteria form a comprehensive instance-level active sampling framework tailored for SAOOD.

\input{algorithm/algorithm1}

\subsection{Instance-Level Classification Uncertainty}
Uncertainty is a widely used measure in active learning. Following prior work \cite{wu2022entropy}, we adopt the information entropy of the predicted class probability distribution from the teacher model as the instance-level classification uncertainty. Specifically, for each object, the teacher model outputs a normalized probability distribution $p = [p_1, p_2, \ldots, p_C]$, where $p_c$ represents the probability that the object belongs to class $c$. The instance-level classification uncertainty is defined as the information entropy:
\begin{equation}
U_{\text{cls}} = - \sum_{c=1}^{C} p_c \log(p_c),
\end{equation}
Higher entropy indicates greater class prediction uncertainty and lower classification confidence.

\subsection{Instance-Level Localization and Orientation Uncertainty}
In oriented object detection, the localization task can be decomposed into two components: bounding box localization and orientation prediction. Since previous active learning methods mainly focus on image-level annotation value, with limited exploration of instance-level localization uncertainty. Moreover, no prior work has explored orientation uncertainty. To this end, we propose a novel instance-level metric for localization and orientation uncertainty. Unlike the classification branch, the localization branch does not output a probabilistic distribution, making it difficult to directly evaluate the accuracy of bounding box. Therefore, we design a plug-and-play localization and orientation uncertainty prediction (LUP) module, which can be integrated into any detector with minimal additional training cost. As shown in Fig. \ref{Fig2}, taking Rotated FCOS as an example, we add a parallel uncertainty prediction branch to the last convolutional layer of the regression head. The uncertainty branch consists of a single $3 \times 3$ convolution with one output channel, responsible for predicting the localization and orientation uncertainty of each rotated bounding box.

During training, the LUP module requires a supervision signal. To better assess the localization uncertainty of a rotated box, we use the Rotated Intersection over Union (RIoU) between the predicted box and the ground truth box as the uncertainty metric, defined as:
\begin{equation}
U_{\text{loc}} = 1 - \text{RIoU}(B_{\text{pred}}, B_{\text{gt}}),
\end{equation}
where $B_{\text{pred}}$ and $B_{\text{gt}}$ denote the predicted box and ground truth, respectively. A larger RIoU corresponds to smaller localization uncertainty, indicating that the object is easy to localize and contributes less to improving the localization capability. Conversely, instances with high localization uncertainty are considered difficult samples, and labeling them can effectively improve the localization performance of model.

Although RIoU can indirectly reflect the orientation accuracy of rotated boxes, it is inherently a strongly coupled measure of both localization and orientation. Previous studies \cite{ARS_DETR} have shown that even for elongated objects with an aspect ratio of 5:1, the RIoU can still remain above 0.5 when the angular deviation is below 15°. This indicates that RIoU alone cannot accurately capture angular errors. Therefore, we further propose orientation uncertainty. Since orientation uncertainty has not been explored in prior work, we introduce a new orientation uncertainty metric. Specifically, we measure the angular deviation between the predicted and ground truth boxes. In oriented object detection, there exists a edge-swapping problem. When the predicted box and the ground truth have different long-side orientations, for example, when the predicted box takes the width as its long side while the ground truth takes the height, they are geometrically equivalent under a $90^\circ$ rotation. Therefore, the predicted angle needs to be corrected to ensure that the long-side direction of the predicted box is consistent with that of the ground truth. To achieve this, we consider edge-swapping problem during the computation of angle deviation. For a predicted box $(x_p, y_p, w_p, h_p, \theta_p)$ and a ground truth box $(x_g, y_g, w_g, h_g, \theta_g)$, where $x$ and $y$ denote the center coordinates, $w$ and $h$ denote the width and height, and $\theta$ is the angle, we first determine whether their long-side directions are consistent:
\begin{equation}
s_p = [w_p \ge h_p], \quad s_g = [w_g \ge h_g],
\end{equation}
where $[\cdot]$ denotes the indicator function. If the long-side directions differ, i.e., $s_p \oplus s_g = 1$ where $\oplus$ represents the XOR operation, the predicted and ground truth boxes are geometrically equivalent under a $90^\circ$ rotation. In this case, we correct the predicted angle as:
\begin{equation}
\theta_p' =
\begin{cases}
\theta_p + \frac{\pi}{2}, & \text{if } s_p \oplus s_g = 1, \\
\theta_p, & \text{otherwise}.
\end{cases}
\end{equation}
After this correction, the long-side orientation of the predicted box becomes consistent with that of the ground truth, ensuring geometric alignment between the two. Then, the angles are further normalized into the range $[-\frac{\pi}{2}, \frac{\pi}{2})$. Since oriented object detection suffers from the angle boundary problem \cite{ACM}, directly computing the angular difference between the predicted box and ground truth may lead to discrepancies between the numerical and actual angular differences, causing instability in the training of the angle uncertainty prediction branch. Therefore, we compute the minimal angular difference as the supervision signal for angle uncertainty:
\begin{equation}
U_{\theta} = \min\big(|{\theta}_p' - {\theta}_g|, \pi - |{\theta}_p' - {\theta}_g|\big),
\end{equation}
where $\theta_p'$ is the corrected predicted angle and $\theta_g$ is the ground truth angle. When the predicted orientation closely matches the ground truth, the orientation uncertainty is low, indicating that these samples are easier for orientation learning. Conversely, samples with high orientation uncertainty pose greater challenges to the orientation estimation. Actively annotating such samples can effectively enhance the orientation prediction ability.

Considering that objects with different aspect ratios have different sensitivity for angular accuracy, where those with larger aspect ratios are more sensitive to angle deviation \cite{zeng2024ars}. Therefore, objects with large aspect ratios should be assigned greater weight on orientation uncertainty. Accordingly, the final instance-level localization and orientation uncertainty is defined as:
\begin{equation}
U_{\mathrm{loc\_\theta}} = w \,U_{\text{loc}} + (1 - w)\,\frac{U_{\theta}}{\pi/2}, \quad w = e^{-\beta \left|\log\!\left(\frac{h_o}{w_o}\right)\right|},
\end{equation}
where $w_o$ and $h_o$ denote the width and height of the object, and $\beta$ is a scaling factor that controls sensitivity to the aspect ratio, which is set to 0.5 in this paper. This formulation adaptively assigns higher orientation uncertainty weights to objects with larger aspect ratios, while assigning higher localization uncertainty weights to near-square objects.

Since the localization and orientation uncertainty is normalized within the range $[0,1]$, the final loss function for the LUP module is formulated as a cross-entropy loss between the predicted uncertainties and the ground truth values:
\begin{equation}
\mathcal{L}_{\text{LUP}} = - \sum \Big[ U_{\mathrm{loc\_\theta}} \log \hat{U}_{\mathrm{loc\_\theta}} + (1 - U_{\mathrm{loc\_\theta}}) \log (1 - \hat{U}_{\mathrm{loc\_\theta}}) \Big],
\end{equation}
where $\hat{U}_{\mathrm{loc\_\theta}}$ denotes the predicted localization and orientation uncertainty for each instance.

\subsection{Instance-Level Inter- and Intra-Class Diversity}
The long-tailed distribution in remote sensing images significantly affects the detection accuracy of rare categories, especially under low sparse annotation rates, where the number of annotated rare instances is extremely limited, exacerbating this issue. To address this, we propose instance-level inter-class diversity to increase the sampling priority of rare categories. We define instance-level inter-class diversity using a global adjustment factor based on class frequency. Specifically, let $N_c$ denote the number of labeled samples for class $c$, the inter-class diversity $D_{\text{inter}}$ is defined as a sigmoid-like function:
\begin{equation}
D_{\text{inter}} = \frac{1}{1 + e^{(\gamma N_c - 1)}},
\end{equation}
where $\gamma$ is the smoothing coefficient, which is set to 0.01 in this work. When a class has fewer samples, $D_\text{inter}$ is higher, encouraging selection of that class to improve overall category diversity. As the number of samples increases, $D_c$ decreases, suppressing further sampling of that class.

To address the large intra-class variations in remote sensing images, we further propose intra-class diversity to ensure that instance features within each class are rich and representative. We first compute the mean classification feature vector of initially annotated samples as the class prototype. During active learning, the cosine similarity between the feature of a candidate instance and its class prototype is used to compute intra-class diversity:
\begin{equation}
D_{\text{intra}} = 1 - \frac{\mathbf{f}_i \cdot \mathbf{P}_c}{\|\mathbf{f}_i\| \, \|\mathbf{P}_c\|},
\end{equation}
where $\mathbf{f}_i$ denotes the classification feature of the $i$-th instance and $\mathbf{P}_{c}$ represents the prototype feature of the class to which the $i$-th instance belongs. A higher $D_{\text{intra}}$ value indicates that the instance contributes novel information to the feature space, helping the model learn more diverse features and improving the generalization.

To dynamically update the prototypes of selected samples for each class, we adopt an exponential moving average (EMA) strategy:
\begin{equation}
\mathbf{P}_c \leftarrow \alpha \mathbf{P}_c + (1-\alpha)\mathbf{f}_i,
\end{equation}
where $\alpha$ is the momentum for intra-class diversity, which is set to 0.9 in this work. This mechanism smooths the prototype updates, reduces noise, and ensures that prototypes gradually reflect the evolving feature distribution of annotated instances.


\subsection{Model State Observation Strategy}
For image-level active learning, it is sufficient to directly select the most difficult images, since each image typically contains a diverse set of instances with varying difficulty. However, for instance-level active learning, selecting a large number of difficult instances at the early stages of training may adversely affect training stability \cite{curriculum_learning}. Curriculum learning studies have shown that gradually training from easy to hard samples is more model-friendly. Moreover, for different instance-level evaluation dimensions, a simple summation to obtain the final sample value may be suboptimal for the current model. To address these issues, we propose a model state observation strategy (MSO).

The module evaluates the current capability of the model across multiple dimensions in a state space and guides the sample selection module to pick high-value samples most suitable for the current model state. Specifically, it observes the performance of model on the training set in terms of classification, localization, inter-class diversity, and intra-class diversity, aligning each dimension with the ultimate objectives of oriented object detection: classification ability is measured by mean Average Precision (mAP) across all classes, localization ability by the mean IoU of predicted boxes matched to the sparse ground truth, inter-class diversity to improve rare-class performance, and intra-class diversity to enhance robustness within each class. Notably, the purpose of our MSO is to observe the relative evolution trends of model ability states across different dimensions during training, rather than to evaluate the absolute mAP of the final model in the conventional sense. These four dimensions are computed as:
\begin{equation}
\begin{aligned}
A_{\mathrm{cls}} &= \frac{1}{C}\sum_{c=1}^{C} AP_c, 
\quad
A_{\mathrm{loc}} = \frac{1}{N}\sum_{i=1}^{N} \mathrm{IoU}\!\left(b_i^{\mathrm{pred}}, b_i^{\mathrm{GT}}\right), \\
A_{\mathrm{inter}} &= \sum_{c \in \mathrm{rare}} AP_c,
\quad
A_{\mathrm{intra}} = \frac{1}{C}\sum_{c=1}^{C} \mathrm{Acc}_c \cdot \exp\!\left(-\mathrm{Var}_c\right),
\end{aligned}
\label{eq11}
\end{equation}
where $C$ is the number of classes, $N$ the number of matched predictions, $AP_c$ the average precision of class $c$, $b_i^\text{pred}$ and $b_i^\text{GT}$ the predicted and GT boxes, $\text{Acc}_c$ the accuracy of class $c$, and $\text{Var}_c$ the confidence variance for class $c$. The final annotation value is computed using a softmax-like composite scoring:
\begin{equation}
S=\sum_{i}
\frac{e^{\,1 - A_i}}{\sum_{j} e^{\,1 - A_j}}
\, U_i,
\label{eq12}
\end{equation}
where $i \in \{\mathrm{cls},\,\mathrm{loc\_\theta},\,\mathrm{inter},\,\mathrm{intra}\}$, and $U_i$ respectively represents classification, localization and orientation uncertainty, as well as inter- and intra-class diversity. The final sample score is then computed as:
\begin{equation}
S_{\mathrm{final}} = \bar{A} \cdot S + (1 - \bar{A}) \cdot (1 - S),
\label{eq:final}
\end{equation}
where $\bar{A}$ denotes the average model capability value across all dimensions. Based on the model capability values measured by the proposed model state observation strategy, higher weights are adaptively assigned to dimensions in which the model is currently weaker during sample selection, ensuring that the selected samples are well matched to the current learning state of model and aligned with the ultimate objectives of oriented object detection. This strategy avoids manually designed heuristics and leverages the own training status of model to guide sample selection, enabling more effective and principled instance selection. In early training stages, when the model capability is limited, Eq. \ref{eq12} favors relatively easy samples, while as training progresses and the model capability gradually improves, increasingly harder samples are selected. This adaptive, state-aware selection process naturally improves training stability and progressively enhances detection performance.

%% file: Figure/img2.tex
\begin{figure*}[t]
    \centering
    \includegraphics[width=0.95\textwidth]{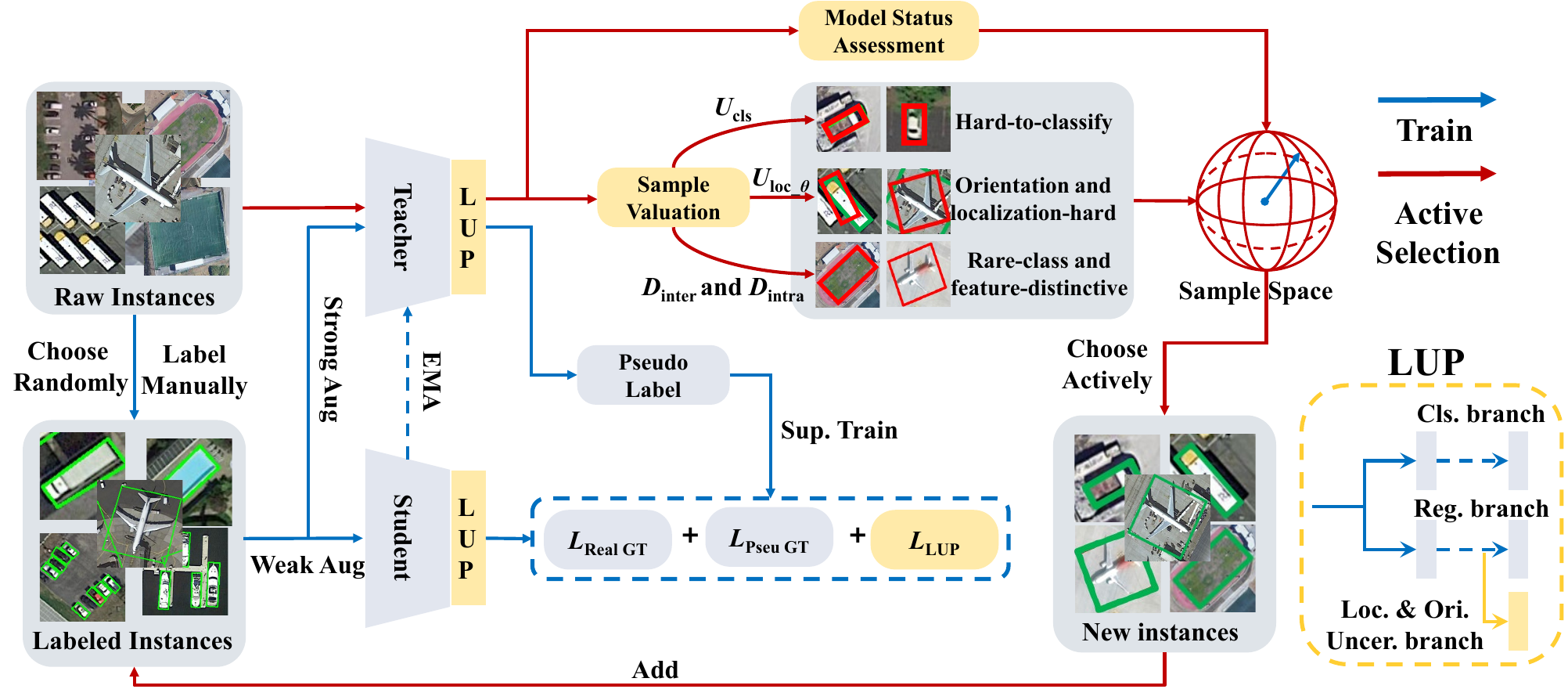}
    \caption{Overall framework of Active-SAOOD. A small subset of training instances is sparsely annotated to initialize the teacher and student models. A model state observation module assesses the capability of teacher across classification, localization, and orientation. Guided by this module, the teacher evaluates unlabeled instances on multiple dimensions, to actively select the most valuable samples, progressively enhancing the training labels.}
    \label{Fig2}
\end{figure*}

%% file: algorithm/algorithm1.tex
\begin{algorithm}[t]
\caption{Pseudo Code of Active-SAOOD}
\label{alg:active-saoood}
\begin{algorithmic}[1]

\REQUIRE Sparsely annotated instance set $\mathcal{D}_I$; 
         Unlabeled instance set $\mathcal{D}_U$; 
         Maximum iteration $N$
\ENSURE Trained oriented object detector $M^t$

\STATE Train initial student model $M^s_0$ on $\mathcal{D}_I$ and $\mathcal{D}_U$
\STATE Initialize teacher model $M^t_0$ via EMA

\FOR{$n = 1, \ldots, N$}
  \STATE Observe model state $\mathcal{O}_{n-1}$ using model state observation module (MSO) based on $M^t_{n-1}$
  \FORALL{$x_u \in \mathcal{D}_U^{n-1}$}
    \STATE Evaluate annotation value $S(x_u \mid \mathcal{O}_{n-1})$ using teacher model $M^t_{n-1}$ via Eq.~(\ref{eq12})
  \ENDFOR
  \STATE Rank all unlabeled instances by $S(x_u \mid \mathcal{O}_{n-1})$
  \STATE Select top-$K$ instances $\mathcal{X}_A^k$ and annotate labels $\mathcal{Y}_A^k$
  \STATE Update labeled set $\mathcal{D}_A^n = \mathcal{D}_A^{n-1} \cup \{\mathcal{X}_A^k, \mathcal{Y}_A^k\}$
  \STATE Update unlabeled set $\mathcal{D}_U^n = \mathcal{D}_U^{n-1} \setminus \{\mathcal{X}_A^k\}$
  \STATE Retrain student model $M^s_n$ on updated $\mathcal{D}_A^n$
  \STATE Update teacher model $M^t_n$ using EMA
\ENDFOR
\STATE \textbf{return} $M^t_N$

\end{algorithmic}

\label{algorithm1}
\end{algorithm}

%% file: sec/4_experiments.tex
\section{Experiments}
\subsection{Dataset and Metric}
To validate our method, we use two oriented object detection datasets: DOTA-v1.0 \cite{DOTA} with 2,806 images and 188,282 instances across 15 categories (cropped to $1024\times1024$ patches) and DIOR \cite{DIOR} with 23,463 images and 192,472 instances in 20 categories. For SAOOD, we create a sparse dataset by randomly selecting $p\%$ of instances without class priors, ensuring at least one instance per category. Evaluation follows prior work \cite{PECL} using mAP$_{50}$.

\subsection{Experimental Settings} 
Following previous work \cite{SOOD}, we use Rotated FCOS \cite{FCOS} with a ResNet-50 backbone pretrained on ImageNet as the baseline detector. Implementation and hyperparameters follow the default mmrotate settings. For active learning, following the previous work \cite{Active_teacher}, algorithm \ref{algorithm1} runs for 2 iterations. Experiments are conducted on four NVIDIA RTX 3090 GPUs with batch size 4. Following common practice \cite{xu2021end}, the EMA momentum in the teacher–student framework is 0.9996, and the LUP module loss weight is 1.
\input{Table/Tab1}
\subsection{Main Results} 
To validate the effectiveness of Active-SAOOD, we conduct experiments on DOTA-v1.0 under different sparse annotation rates. As shown in Table \ref{Tab1}, compared with random annotation, Active-SAOOD evaluates samples across multiple instance-level dimensions and, guided by the model state observation module, selects instances most suitable for the current model state. This ensures selected samples are valuable and aligned with model capability. Notably, at an extremely low annotation rate (1\%), the baseline with random annotation achieves 20.94\% mAP, while using our actively selected high-value samples improved it to 30.24\%. For SAOOD methods, Active-SAOOD consistently boosts performance: using RSST, random annotation reaches 23.19\% mAP, whereas Active-SAOOD (RSST-based) attains 38.54\%. For S$^2$Teacher, random annotation obtains 30.73\%, while Active-SAOOD reaches 35.79\%. Random annotation easily selects less informative instances, leading to lower performance and unstable training, whereas Active-SAOOD prioritizes high-value samples, making better use of annotation budgets and improving both performance and stability. When samples are selected based on image-level evaluation dimensions, such as Active Teacher or PPAL, while pseudo-labels are mined with S$^2$Teacher, the mAP under the 1\% annotation ratio is only 29.2\% and 30.71\%, respectively. In contrast, our instance-level Active-SAOOD achieves 35.79\%. This is mainly because conventional image-level active learning evaluation dimensions and selection strategies are not well suited for the sparse annotation setting, resulting in selected samples that are less compatible with the model state. These results highlight the advantage of selection guided by MSO.

\subsection{Generalization Experiment on DIOR}
To evaluate generalization, we conducted experiments on the DIOR dataset. As Table \ref{Tab2} shows, Active-SAOOD consistently improves SAOOD performance under different sparse annotation ratios. Compared with DOTA, DIOR dataset has more object categories but fewer instances, making valuable sample selection harder. At a 1\% annotation rate, random annotation achieved only 15.64\% mAP, while Active-SAOOD reached 25.58\%, demonstrating its strong generalization.
\input{Table/Tab2}
\input{Figure/img3}
\subsection{Ablation Study}
To verify the effectiveness of the proposed instance-level active learning evaluation metrics, we conducted ablation experiments on DOTA-v1.0 with 1\% annotations. As shown in Table \ref{Tab3}, removing any of the four evaluation dimensions, including classification uncertainty, localization and orientation uncertainty, inter-class diversity, and intra-class diversity, leads to varying degrees of performance degradation. This is because these four dimensions are specifically designed based on the intrinsic characteristics of the SAOOD task, and each dimension influences the value of selected samples from a different perspective, thereby affecting the final detection performance. Classification uncertainty improves the classification capability by selecting samples with higher classification difficulty. Localization and orientation uncertainty focuses on samples that are difficult to localize or estimate the angle, enhancing the localization and orientation ability and thereby improving overall detection performance. Inter-class diversity is designed to address the long-tailed distribution in remote sensing images, helping to balance the number of annotated instances across different categories and to mitigate class bias during the prediction process. Finally, intra-class diversity is introduced to address the large intra-class variations in remote sensing images, preventing the selection of overly feature-similar samples that may cause overfitting, thereby enhancing the intra-class robustness of model. MSO observes the capabilities of model in the state space and guides sample selection module to adaptively choose samples suited to the current model state, which further improves performance.
\input{Table/Tab3}
\input{Table/Tab4}
\subsection{Repeat Annotation Experiment}
To demonstrate that SAOOD improves stability compared with random sparse annotation, we conducted repeated experiments on DOTA with a 1\% annotation rate. As shown in Table \ref{Tab4}, over five runs, random annotation produced highly variable results, with mAP ranging from 4.42\% to 24.03\% and a variance of 74.12, sometimes even causing training failure. SAOOD method with pseudo-label mining reached a maximum of 30.73\% mAP, but variance remained high at 62.59. In contrast, Active-SAOOD achieved a maximum of 35.79\% and a minimum of 30.46\%, with a variance of only 4.08, improving performance by 11.76 and 26.04\% over random annotation. These results demonstrate that Active-SAOOD can maintain stable, high performance under extremely sparse annotation by actively selecting the most valuable samples. This stability is crucial in real-world annotation, where random annotation failures can significantly raise labeling costs and hinder practical use of SAOOD.
\subsection{Visualization Analysis}
To better understand the samples selected by different criteria, we conducted a visualization analysis. As illustrated in Fig. \ref{Fig3}, samples selected based on classification uncertainty tend to correspond to instances with ambiguous visual features and higher classification difficulty. Samples chosen under localization uncertainty are typically those whose predicted boxes exhibit larger positional deviations, indicating higher localization difficulty. When orientation uncertainty is considered, the selected samples are those with greater errors in angle prediction, representing more challenging oriented instances. Incorporating inter-class diversity encourages the selection of instances from rare categories, while intra-class diversity favors the selection of visually distinctive samples within each class, enriching the diversity of features. These visualizations demonstrate that Active-SAOOD effectively selects the most valuable instances for annotation from multiple perspectives, including orientation, classification, and localization uncertainty, as well as inter- and intra-class diversity, thereby enhancing the informational richness and ensuring model performance.

%% file: Table/Tab1.tex
\begin{table*}[t]
  \centering
  \caption{Comparison of our Active-SAOOD with other SAOOD and image-level active learning (AL) methods on DOTA-v1.0 under different annotation rate. mAP is used for evaluation, and $\Delta$ denotes the mAP gain over the baseline. Active-Rotated FCOS is trained only with actively selected high-value samples, while Active-SAOOD additionally incorporates mined pseudo-labels.}
  \resizebox{0.9\textwidth}{!}{
  \begin{tabular}{ccccccccccc}
    \toprule
    \multirow{2}[4]{*}{Method paradigm} & \multirow{2}[4]{*}{Method} & \multicolumn{8}{c}{Sparse annotation rate} \\
    \cmidrule{3-10}
    & & 1\% & $\Delta$ & 2\% & $\Delta$ & 5\% & $\Delta$ & 10\% & $\Delta$ \\
    \midrule
    Baseline & Rotated FCOS & 20.94 & +0.00 & 33.23 & +0.00 & 48.59 & +0.00 & 55.06 & +0.00 \\
    \midrule
    \multirow{3}{*}{SAOOD} &
    DML \cite{DML} & 21.62 & +0.68 & 34.91 & +1.68 & 52.51 & +1.92 & 59.16 & +4.10 \\
    & RSST \cite{RSST} & 23.19 & +2.25 & 37.20 & +3.97 & 55.22 & +6.63 & 60.62 & +5.56 \\
    & S$^2$Teacher \cite{S2Teacher} & 30.73 & +9.79 & 37.86 & +4.63 & 55.49 & +6.90 & 61.67 & +6.61 \\
    \midrule
    \multirow{2}{*}{Image-level AL} &
    Active Teacher \cite{Active_teacher} & 29.20 & +8.26 & 37.55 & +4.32 & 55.06 & +6.47 & 60.72 & +5.66 \\
    & PPAL \cite{PPAL} & 30.71 & +9.77 & 37.62 & +4.39 & 56.23 & +7.64 & 60.75 & +5.69 \\
    \midrule
    \multirow{4}{*}{Instance-level AL} &
    Active-Rotated FCOS & 30.24 & +9.30 & 35.42 & +2.19 & 53.89 & +5.30 & 57.90 & +2.84 \\
    & Active-SAOOD (DML-based) & 32.24 & +11.30 & 40.49 & +7.26 & 56.12 & +7.53 & 60.12 & +5.06 \\
    & Active-SAOOD (RSST-based) & \textbf{38.54} & +17.6 & \textbf{43.09} & +9.86 & 57.99 & +9.40 & 61.16 & +6.10 \\
    & Active-SAOOD (S$^2$Teacher-based) & 35.79 & +14.85 & 42.98 & +9.75 & \textbf{58.39} & +9.80 & \textbf{62.61} & +7.55 \\
    \bottomrule
  \end{tabular}
  }
  \label{Tab1}
\end{table*}

%% file: Table/Tab2.tex
\begin{table}[t]
  \centering
  \caption{Generalization experiments on DIOR.}
  \resizebox{0.9\linewidth}{!}{
  \begin{tabular}{cccc}
    \toprule
    Annotation Rate & Method & mAP (\%) & $\Delta$ \\
    \midrule
    \multirow{3}[2]{*}{1\%}
      & Rotated FCOS & 15.64 & +0.00 \\
      & SAOOD \cite{S2Teacher} & 23.17 & +7.53 \\
      & Active-SAOOD & \textbf{25.58} & +9.94 \\
    \midrule
    \multirow{3}[2]{*}{2\%}
      & Rotated FCOS & 17.89 & +0.00 \\
      & SAOOD & 20.57 & +2.68 \\
      & Active-SAOOD & \textbf{32.73} & +14.84 \\
    \midrule
    \multirow{3}[2]{*}{5\%}
      & Rotated FCOS & 33.41 & +0.00 \\
      & SAOOD & 37.06 & +3.65 \\
      & Active-SAOOD & 4\textbf{0.73} & +7.32 \\
    \midrule
    \multirow{3}[2]{*}{10\%}
      & Rotated FCOS & 44.83 & +0.00 \\
      & SAOOD & 48.79 & +3.96 \\
      & Active-SAOOD & \textbf{50.84} & +6.01 \\
    \bottomrule
  \end{tabular}
  }
  \label{Tab2}
\end{table}

%% file: Figure/img3.tex
\begin{figure*}[t]
    \centering
    \begin{subfigure}{0.95\textwidth}
        \centering
        \includegraphics[width=\textwidth]{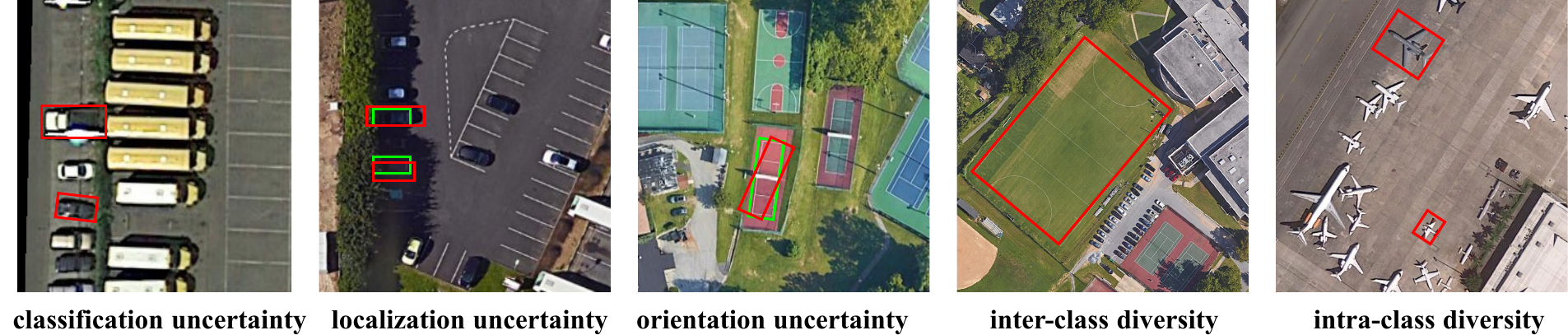}
        \caption{}
        \label{fig:2a}
    \end{subfigure}

    \vspace{1em} 

    \begin{subfigure}{0.95\textwidth}
        \centering
        \includegraphics[width=\textwidth]{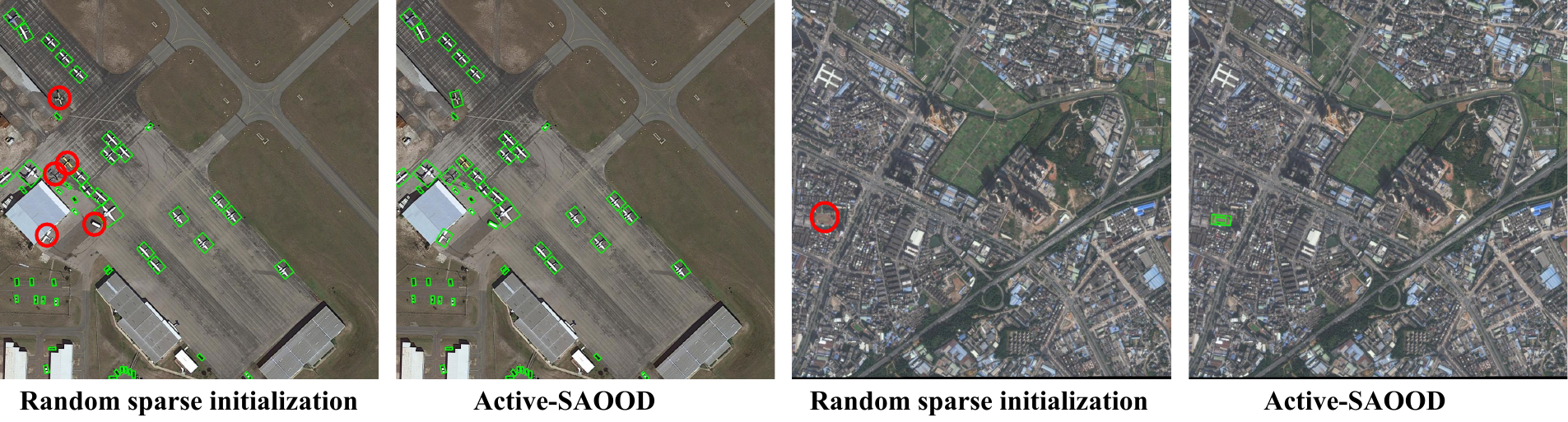}
        \caption{}
        \label{fig:2b}
    \end{subfigure}

    \caption{Visualization of instance selection and model performance. (a) Visualization of the actively selected instances. (b) Comparison of detection results for models trained on random versus actively selected instances.}
    \label{Fig3}
    
\end{figure*}

%% file: Table/Tab3.tex
\begin{table}[t]
  \centering
  \caption{Ablation study of different evaluation dimensions on DOTA-v1.0 with 1\% annotation rate.}
  \resizebox{0.8\linewidth}{!}{
  \begin{tabular}{cccccc}
    \toprule
    $\mathrm{U_{cls}}$ & $\mathrm{U_{loc\_\theta}}$ & $\mathrm{D_{inter}}$ & $\mathrm{D_{intra}}$ & MSO & mAP(\%) \\
    \midrule
    \checkmark & \checkmark & \checkmark & \checkmark & \checkmark & \textbf{35.79} \\
    -- & \checkmark & \checkmark & \checkmark & \checkmark & 33.62 \\
    \checkmark & -- & \checkmark & \checkmark & \checkmark & 34.75 \\
    \checkmark & \checkmark & -- & \checkmark & \checkmark & 32.66 \\
    \checkmark & \checkmark & \checkmark & -- & \checkmark & 33.90 \\
    \checkmark & \checkmark & \checkmark & \checkmark & -- & 32.70 \\
    \bottomrule
  \end{tabular}
  }
  \label{Tab3}
\end{table}

%% file: Table/Tab4.tex
\begin{table}[t]
  \centering
  \caption{Repeated experiments on DOTA-v1.0 with 1\% sparsely annotation rate. The mAP is the evaluation metric in the table, and var denotes the variance.}
  \resizebox{0.9\linewidth}{!}{
    \begin{tabular}{ccccc}
    \toprule
    Method & max   & min   & mean  & var \\
    \midrule
    Random annotation & 24.03 & 4.42  & 16.46 & 74.12 \\
    SAOOD & 30.73 & 12.15 & 26.25 & 62.59 \\
    Active-FCOS & 30.24 & 25.52 & 27.35 & 4.29 \\
    Active-SAOOD & \textbf{35.79} & \textbf{30.46} & \textbf{32.80} & \textbf{4.08} \\
    \bottomrule
    \end{tabular}%
  }
  \label{Tab4}
  
\end{table}

%% file: sec/5_conclusion.tex
\section{Conclusion}
In this work, we propose Active-SAOOD, an instance-level active learning framework to address the performance degradation and instability of SAOOD under random sparse annotation. By jointly considering instance-level classification, localization, and orientation uncertainties as well as inter- and intra-class diversity, Active-SAOOD selects high-value samples guided by a model state observation strategy. Experimental results show that Active-SAOOD significantly outperforms random annotation and improves the stability of SAOOD under low sparse annotation rates, facilitating broader practical applications.

%% file: sec/X_suppl.tex
\section{More detailed experimental settings.}
Following most low-cost annotated oriented object detection studies \cite{H2RBox, SOOD}, we adopt Rotated FCOS as our baseline detector, with a ResNet-50 backbone pretrained on ImageNet. Focal Loss is used for classification, with its parameters configured according to \cite{Focal}. IoU Loss is employed for regression, and a cross-entropy loss is applied to centerness. All loss weights are set to 1. Consistent with prior work \cite{S2Teacher, lin2026robust}, we use a teacher–student framework to mine pseudo labels for sparse-annotation training in oriented object detection. Following common practice \cite{SOOD, lin2024weakly}, the EMA momentum of the teacher model is set to 0.9996, and the teacher parameters are updated via EMA at each iteration \cite{lin2025you}. All remaining settings follow the standard configurations in mmrotate. To ensure a fair comparison, we strictly match the number of annotated objects between the randomly sparse annotations and those actively selected by our Active-SAOOD approach.

\section{More detailed experiments.}
\subsection{More detailed hyperparameter experiments.}
To analyze how the hyperparameters in our evaluation metrics influence both the selected instances and the final detection performance, we conduct detailed hyperparameter experiments. The hyperparameter $\gamma$ in the inter-class diversity controls the slope of the sigmoid-like function. A larger $\gamma$ yields a steeper slope, making the inter-class diversity more sensitive to the sampled number $N_c$. However, an excessively large $\gamma$ causes the diversity score to drop too rapidly, weakening the contribution of this metric, while an overly small $\gamma$ makes the diversity score insensitive to $N_c$, limiting its ability to highlight rare categories. As shown in Table \ref{Tab_supp1}, both overly large and overly small $\gamma$ values degrade the effectiveness of the inter-class diversity, ultimately leading to performance drops in Active-SAOOD. When $\gamma$ is set too small (e.g., $0.001$), the inter-class diversity score is excessively large, which can mislead the overall evaluation of instance annotation value and result in training failure. When $\gamma$ is set to $0.01$, Active-SAOOD achieves the best performance, indicating that this value provides the most effective balance. Therefore, we set $\gamma=0.01$ in experiments. We also conduct hyperparameter studies on $\beta$ in the localization and orientation uncertainty metric. $\beta$ controls the relative emphasis between orientation and localization uncertainties. In the extreme case where $\beta=0$, $U_{\text{loc}\_\theta}$ is determined solely by localization uncertainty. As $\beta$ increases, the metric increasingly prioritizes orientation uncertainty, particularly for objects with large aspect ratios. When $\beta$ becomes excessively large, the orientation uncertainty for high--aspect-ratio objects dominates the metric, causing $U_{\text{loc}\_\theta}$ to overlook localization uncertainty, which adversely affects performance. As shown in Table \ref{Tab_supp1}, setting $\beta=0.5$ yields the highest performance of Active-SAOOD. Therefore, we use $\beta=0.5$ in experiments.
\input{Table/Tab5}

\subsection{Ablation study on evaluation dimensions and model abilities.}
To further demonstrate that samples selected by different evaluation dimensions indeed affect the model capability along corresponding aspects during training, we conduct a more fine-grained ablation study. As shown in Table \ref{Tab_add1}, when the classification uncertainty dimension is removed during sample selection, the classification accuracy of the trained model drops to 76.94\%, representing a decrease of 3.66\% compared with the full model. This is mainly because classification uncertainty emphasizes selecting hard-to-classify instances to enhance the classification capability of model. Without this evaluation dimension, the actively selected samples lack sufficient classification difficulty, leading to degraded classification performance. Localization uncertainty focuses on selecting instances that are difficult to localize, thereby improving the localization capability of model. Removing this dimension results in weaker localization performance, which is reflected by a decrease of 10.65\% in the average IoU. Inter-class diversity prioritizes instances from rare categories to encourage a more balanced class distribution. This strategy effectively alleviates the long-tailed distribution issue commonly observed in remote sensing images, where rare categories tend to suffer from inferior performance. When inter-class diversity is removed, the classification accuracy of rare categories decreases by 3.34\%. Intra-class diversity aims to select instances with more distinctive features, which enhances intra-class robustness and helps address the large intra-class variations in remote sensing images. Removing intra-class diversity leads to a decline in the mean recall across categories, indicating reduced detection robustness for visually distinctive instances. Overall, the ablation results and the analysis of model performance across different capability dimensions validate the effectiveness of the proposed instance-level evaluation dimensions, which are designed by explicitly considering the intrinsic characteristics of the SAOOD.


\begin{table}[t]
  \centering
  \caption{Ablation study of evaluation dimensions on model capability. All denotes the full model with all dimensions, and other rows show performance after removing each dimension. Bold values indicate significant drop in the corresponding capability.}
  \resizebox{\linewidth}{!}{
    \begin{tabular}{ccccc}
    \toprule
    Setting & Acc (\%) & IoU (\%) & Rare Acc (\%) & Recall (\%) \\
    \midrule
    All   & 80.60 & 69.77 & 69.86 & 60.30 \\
    $\mathrm{U_{cls}}$  & \textbf{76.94} & 69.06 & 68.56 & 59.26 \\
    $\mathrm{U_{loc\_\theta}}$ & 80.52 & \textbf{59.12} & 69.51 & 60.20 \\
    $\mathrm{D_{inter}}$ & 79.25 & 69.12 & \textbf{66.52} & 59.69 \\
    $\mathrm{D_{intra}}$ & 80.16 & 69.24 & 69.22 & \textbf{55.68} \\
    \bottomrule
    \end{tabular}%
    }
  \label{Tab_add1}%
\end{table}%

\input{Figure/img_add3}

\begin{table}[t]
  \centering
  \caption{Generalization experiments of the MSO strategy. MSO is employed to evaluate the model state on both the initially annotated GT and an independent held-out set of comparable size, respectively, to guide high-value label selection.}
  \resizebox{0.95\linewidth}{!}{
    \begin{tabular}{ccccc}
    \toprule
    \multirow{2}[4]{*}{Ability evalutation setting} & \multicolumn{4}{c}{Sparse annotation rate} \\
\cmidrule{2-5}          & 1\%   & 2\%   & 5\%   & 10\% \\
    \midrule
    Evaluated on initially labeled GT & 35.79 & 42.98 & 58.39 & 62.61 \\
    Evaluated on held-out set & 34.96 & 42.67 & 58.32 & 62.60  \\
    \bottomrule
    \end{tabular}%
  \label{Tab_add3}%
  }
\end{table}%
\begin{table*}[t]
  \centering
  \caption{Experiments on the evolution trends of model ability across different training stages. Average ability is the mean of model ability dimensions in Eq. \ref{eq11}. The MSO strategy is employed to evaluate the current model ability on both the initially annotated GT set involved in training and the evaluation data excluded from training.}
  \resizebox{0.95\linewidth}{!}{
    \begin{tabular}{ccccccc}
    \toprule
    \multirow{2}[4]{*}{Annotation rate} & \multirow{2}[4]{*}{Evaluation Set} & \multirow{2}[4]{*}{Metric} & \multicolumn{4}{c}{Train stage} \\
\cmidrule{4-7}          &       &       & Early stage & Middle stage & Late stage & Convergence stage \\
    \midrule
    \multirow{2}[2]{*}{1\%} & Initially labeled GT & \multirow{2}[2]{*}{Average Ability} & 0.49  & 0.63 (+0.14) & 0.702 (+0.072) & 0.705 (+0.003) \\
          & Non-training evaluation set &       & 0.40  & 0.49 (+0.09) & 0.532 (+0.042) & 0.534 (+0.002) \\
    \midrule
    \multirow{2}[2]{*}{5\%} & Initially labeled GT & \multirow{2}[2]{*}{Average Ability} & 0.60  & 0.72 (+0.12) & 0.750 (+0.030) & 0.751 (+0.002) \\
          & Non-training evaluation set &       & 0.56  & 0.67 (+0.11) & 0.698 (+0.028) & 0.699 (+0.001) \\
    \bottomrule
    \end{tabular}%
  \label{Tab_add4}%
  }
\end{table*}%

\subsection{Generalization experiments of the MSO model state observation.}
Our MSO strategy guides Active-SAOOD in selecting the most suitable high-value samples by observing the model state on the initially annotated ground-truth (GT). To verify its generalization ability, we conduct corresponding experiments. To avoid data leakage caused by introducing validation or test data during training, the model state cannot be evaluated on validation or test sets to guide sample selection. Therefore, we study two settings for model state observation using MSO: (1) observated on the initially annotated GT set, and (2) observated on an independently held-out subset of comparable size that is randomly sampled and non-overlapping with the initially annotated GT set. In both cases, the observed model state is used to guide the selection of high-value instances for subsequent training.

As shown in Table \ref{Tab_add3}, the final performance of Active-SAOOD is highly consistent regardless (less than 1\% difference) of whether the model state is evaluated on the initial GT set or on the independent held-out set, and both settings yield significant performance improvements. This indicates that MSO is insensitive to the choice of data used as labels for model state evaluation, as MSO is designed to observe the relative evolution trend of the model’s ability across different dimensions, rather than to estimate the absolute mAP in the conventional sense. Since this evolution reflects the intrinsic state of the model, its trend remains consistent when measured on either the training-participating initial GT or the non-training independent held-out set, leading to nearly identical final performance. Notably, under low annotation ratios (e.g., 1\% annotation ratio), observing the model state on the initially annotated GT set even achieves slightly better performance than using an independent held-out set. This is mainly because, under extremely limited annotations, the model is insufficiently trained, and model state observation on a completely unseen independent held-out set may not accurately reflect the current model state. Consequently, the guided sample selection may become less suitable for the current model state. In addition, annotating an additional independent held-out set would incur extra labeling costs, we ultimately adopt the MSO on the initially annotated GT set for model state observation.

To further validate this observation, we additionally evaluate the model ability on the evaluation set that is not involved in training. As shown in Table \ref{Tab_add4}, the evolution trends of model ability observed on the initially labeled GT are consistent with those observed on non-training evaluation set. Specifically, during the early stage, the model capabilities observed on both are relatively weak, causing MSO to favor relatively easy samples. As training progresses, the model capabilities on both rapidly improve, leading MSO to gradually select challenging samples, and eventually converge in the late training stage. These results indicate that the sample selection guided by MSO based on the initially labeled GT is nearly identical to that guided by observations on non-training data, thereby demonstrating the generalization and rationality of MSO.

\begin{table}[t]
  \centering
  \caption{Generalization experiments on different datasets.}
  \resizebox{0.95\linewidth}{!}{
    \begin{tabular}{cccc}
    \toprule
    \multirow{2}[4]{*}{Dataset} & \multirow{2}[4]{*}{Method} & \multicolumn{2}{c}{Annotation Rate} \\
\cmidrule{3-4}          &       & 1\%   & 5\% \\
    \midrule
    \multirow{3}[2]{*}{DOTA-v1.5} & Rotated FCOS & 21.91 & 42.12 \\
          & S$^2$Teacher & 27.61 & 47.72 \\
          & Active-SAOOD (S$^2$Teacher-based) & \textbf{33.72} & \textbf{50.76} \\
    \midrule
    \multirow{3}[2]{*}{SODA-A} & Rotated FCOS & 37.54 & 59.56 \\
          & S$^2$Teacher & 41.27 & 62.68 \\
          & Active-SAOOD (S$^2$Teacher-based) & \textbf{47.27} & \textbf{66.24} \\
    \bottomrule
    \end{tabular}%
  \label{Tab_add7}%
  }
\end{table}%
\subsection{Experiments on instance difficulty selection guided by MSO.}
To demonstrate the MSO strategy is able to select instances with appropriate difficulty according to the model capability at different training stages, we conduct the experiments. As shown in Table \ref{Tab_add4}, at the early stage of training, the MSO strategy observes that the capabilities of model are relatively weak, and selects relatively simple instances. As training progresses, the capabilities of model rapidly improve, MSO correspondingly observes this enhancement and selects instances of increasing difficulty at the middle stage. At the late stage, MSO observes that the capabilities of model are already strong, and thus selects more difficult instances. Learning from these more challenging samples further boosts performance. Guided by MSO, selecting instances that are most suitable for the current model state, while gradually increasing the difficulty of the selected instances as the ability of model improves, effectively enhances the training stability and the overall performance.

As shown in Fig. \ref{Fig_add3}, we visualize and compare the training loss of the traditional image-level active learning method, which directly selects the most difficult samples, with our state observation–based instance-level active learning approach that gradually increases the sample difficulty according to the model state. The traditional image-level active learning strategy tends to introduce a large number of hard samples at an early stage, which often leads to loss oscillation and unstable training in SAOOD. In contrast, the proposed model state observation–guided active learning strategy progressively selects more difficult instances based on the current model state, thereby ensuring more stable training and superior performance.

\subsection{Generalization experiments on more datasets.}
We further conduct generalization experiments of Active-SAOOD on the DOTA-v1.5 and SODA-A datasets, which contain large numbers of dense scenes and small objects. As shown in Table \ref{Tab_add7}, Active-SAOOD consistently achieves significant performance improvements over S$^2$Teacher under different annotation ratios on both datasets, demonstrating the strong generalization capability of our method in dense scenarios.

Notably, the performance gains become even more significant under the extremely low annotation ratio of 1\%. This is mainly because, under severely limited annotation budgets, random annotation is more likely to produce samples that are unsuitable for effective model learning, thereby wasting valuable annotation costs. In contrast, Active-SAOOD can actively select the most suitable high-value samples that best match the current model state, leading to substantial performance improvements under extremely low annotation ratios.

\begin{table}[t]
  \centering
  \caption{Comparison experiments with other reducing annotation cost methods.}
  \resizebox{\linewidth}{!}{
    \begin{tabular}{cccc}
    \toprule
    Setting & Method & Annotation budget & mAP(\%) \\
    \midrule
    \multirow{3}[2]{*}{Weakly supervised} & H2RBox-v2 & 150\% Point & 72.3 \\
          & Point2RBox-v2 & 100\% Point & 62.61 \\
          & Wholly-WOOD & 100\% Point & 62.63 \\
    \midrule
    Partial weakly supervised & PWOOD & 10\% Point & 42.35 \\
    \midrule
    Semi-supervised & SOOD++ & 20.5\% Point & 54.2 \\
    \midrule
    \multirow{2}[2]{*}{Active learning} & \multirow{2}[2]{*}{Active-SAOOD} & 10.24\% Point & 58.39 \\
          &       & 20.5\% Point & 62.6 \\
    \bottomrule
    \end{tabular}%
  \label{Tab_add5}%
  }
  \vspace{-15pt}
\end{table}%
\subsection{Comparison experiments with other reducing annotation cost methods.}
We further conduct comparison experiments between Active-SAOOD and other reducing annotation costs methods in oriented object detection. Since different settings adopt different annotation forms, we follow the relative annotation costs among RBox, HBox, and Point annotations provided in Wholly-WOOD \cite{Wholly_WOOD} and convert all annotation budgets into an equivalent Point-based budget for a fair comparison. As shown in Table \ref{Tab_add5}, under a 10\% Equivalent point budget, Active-SAOOD reaches 58.39\% AP, significantly outperforming partial weakly-supervised PWOOD \cite{PWOOD} (42.35\%). Moreover, compared with the Point2RBox-v2 \cite{Point2RBox_v2}, Active-SAOOD achieves comparable performance with only one-fifth of the annotation budget. Although H2RBox-v2 \cite{H2RBox_v2} (HBox supervised) achieves higher performance, its annotation budget is much higher than Active-SAOOD. Compared with weakly supervised methods, Active-SAOOD can achieve strong performance under extremely low annotation budgets, with Table \ref{Tab1} highlight that Active-SAOOD achieves greater performance gains at lower annotation budgets, e.g., a 14.85\% gain at 1\% annotation rate and a 9.8\% gain at 5\%. Compared with the semi-supervised method SOOD++ \cite{sood_p}, Active-SAOOD achieves significantly higher performance under the same annotation budget (62.61\% vs. 54.2\%). Unlike weakly- or semi-supervised methods, which cannot select representative instances and suffer from feature redundancy and wasted annotation budget. The key advantage of Active-SAOOD lies in addressing this problem from a novel instance-level active learning perspective, enabling the active selection of high-value and representative instances, thereby achieving comparable or superior performance under extremely low annotation costs.

\begin{table}[t]
  \centering
  \caption{Experiment on the number of active learning selection.}
  \resizebox{0.9\linewidth}{!}{
    \begin{tabular}{ccc}
    \toprule
    Method & Iteration number & mAP (\%) \\
    \midrule
    \multirow{2}[2]{*}{Active-Rotated FCOS} & 2     & 30.24 \\
          & 3     & 30.73 \\
    \midrule
    \multirow{2}[2]{*}{Active-SAOOD} & 2     & 35.79 \\
          & 3     & 37.83 \\
    \bottomrule
    \end{tabular}%
  \label{Tab_add3}%
  }
  \vspace{-15pt}
\end{table}%

\input{Figure/img_add1}
\input{Figure/img_add2}

\subsection{Experiment on the number of active learning iterations.}
We further investigate the effect of the number of selection iterations in the active learning. As shown in Table \ref{Tab_add3}, we set the number of iterations in Algorithm \ref{algorithm1} to different values. When the number of active learning iterations increases from 2 to 3, the performance of Active-Rotated FCOS improves by 0.49\%, while the performance of Active-SAOOD increases from 35.79\% to 37.83\%, yielding an improvement of 2.04\%. This is mainly because a larger number of iterations corresponds to more active selection rounds. For example, increasing the iterations from 2 to 3 introduces an additional intermediate selection step. This allows Active-SAOOD to select more suitable samples based on the model state at intermediate training stages, thereby progressively enhancing detection performance. However, increasing the number of iterations also leads to higher training costs. Considering the trade-off between training cost and detection performance, we recommend setting the number of iterations in Algorithm \ref{algorithm1} to 2.

\begin{table}[t]
  \centering
  \caption{Computational cost experiments of the LUP and MSO.}
  \resizebox{0.95\linewidth}{!}{
    \begin{tabular}{cccc}
    \toprule
    \multirow{2}{*}{Method} & Train time & Parameters & Inference time \\
     & (min) & (M) & (s) \\
    \midrule
    + LUP  & 2  & 0.0046 & 0 \\
    + MSO  & 32 & 0      & 0 \\
    + LUP + MSO & 34 & 0.0046 & 0 \\
    \bottomrule
    \end{tabular}%
  }
  \label{Tab_add6}%
  \vspace{-15pt}
\end{table}
\subsection{Computational cost experiments of LUP and MSO.}
Since our main goal is to reduce annotation costs while improving performance, we conduct a simple analysis of the additional computational costs introduced by LUP and MSO. As shown in Table \ref{Tab_add6}, both LUP and MSO are only applied during training and can be removed during inference, thus they do not affect inference speed. LUP adopts a simple 3 $\times$ 3 convolution, while MSO only involves evaluation metrics and softmax operations, resulting in negligible overhead in both parameters and training time. These results demonstrate that Active-SAOOD does not introduce significant additional training time.

\subsection{Failure cases analysis.}
To gain a deeper understanding of Active-SAOOD, we further conduct a visualization analysis of several failure cases during detection. As shown in Fig. \ref{Fig_add2}, for small vehicles heavily occluded by shadows, the extremely blurred appearance still leads to missed detections. However, we observe that even under such extreme conditions, one small vehicle is still correctly detected. This is mainly because the classification uncertainty encourages the selection of classification-challenging instances for training, which in turn helps the model better recognize such difficult objects. For extremely rare categories such as helicopter, although the inter-class diversity prioritizes selecting these instances for annotation, the number of rare-category samples in the training set remains limited under very low annotation ratios. As a result, some rare objects may still be missed. Nevertheless, compared with random sparse annotation, the detection accuracy for rare categories is already significantly improved. As shown in Fig. \ref{Fig_add2}, Active-SAOOD occasionally misclassifies containers as large vehicles, likely because the increased intra-class diversity enriches the feature space of the training set, and the visual similarity between containers and large vehicles is very high, making them harder to distinguish. In addition, during visualization, we observed that some small vehicles partially occluded in the images were overlooked during the manual annotation. Owing to the improved generalization ability brought by actively selected instances, Active-SAOOD is still able to detect these manually missed small vehicles, which further demonstrates the strong generalization ability of our Active-SAOOD.

\section{More detailed visualization analysis.}
To deeper insights into the high-value instances selected by Active-SAOOD, we conduct a comparative visualization analysis between randomly selected instances and those actively selected by our method. As shown in Fig. \ref{Fig_add1_a}, random selection exhibits strong randomness, and under low annotation ratios it often selects instances with highly similar characteristics, thereby reducing the overall feature diversity of the annotated data. For example, in the left figure, most selected planes share very similar appearances, while more visually distinctive airplanes remain unannotated. In contrast, Active-SAOOD leverages our proposed intra-class diversity metric to evaluate and select instances with more distinctive characteristics, ensuring richer feature diversity even under low annotation ratios. Moreover, compared with random selection, Active-SAOOD tends to actively choose objects with large aspect ratios (such as the large-vehicle in the middle figure). This is because orientation prediction errors have a stronger impact on the localization of such objects, and our localization and orientation uncertainty metric encourages selecting these challenging instances for training. For rare categories (such as the runway and sports field in the right figure), random selection often fails to include them under low annotation ratios. Active-SAOOD, however, prioritizes these rare categories through our inter-class diversity metric, preventing model overfitting and improving detection accuracy for infrequent classes.

To understand how training with actively selected instances benefits the detector, we compare the detection results of models trained on different instance selection strategies. As shown in Fig. \ref{Fig_add1_b}, Rotated FCOS represents the baseline trained with randomly selected sparse annotations, while Active-FCOS denotes the same baseline trained with instances actively selected by our method (with the annotation budget kept identical). Both models share the same network architecture and training settings. SAOOD is trained on randomly selected sparse annotations but incorporates pseudo-label mining, whereas Active-SAOOD uses actively selected instances together with pseudo-label training.

Due to sparse supervision, Rotated FCOS lacks sufficient positive samples, especially under low annotation ratios. The limited diversity of the annotated instances causes the model to overfit easily, resulting in a small number of detected objects. In contrast, Active-FCOS benefits from our actively selected high-value samples, which provide richer feature diversity even under low annotation ratios. This alleviates overfitting to some extent, enabling the model to detect more objects. However, without pseudo-label training, unlabeled objects still confusion foreground and background features, leading to missed detections, particularly for small vehicles. SAOOD alleviates overfitting by mining pseudo-labels for unlabeled objects, thereby increasing the number of positive samples and reducing missed detections. Nevertheless, the localization and orientation accuracy of pseudo-labels is not always reliable. This can negatively impact the ability of model to localize and orient objects accurately, causing angle prediction errors for large aspect-ratio objects (e.g., large-vehicle). Active-SAOOD addresses this issue by selecting localization and orientation challenging objects through our proposed uncertainty dimension, allowing the model to better learn accurate localization and orientation for difficult cases. As a result, large aspect-ratio objects with challenging rotation angles exhibit more precise orientation predictions. For example, the yellow large-vehicle in Fig. \ref{Fig_add1_b} shows notably improved angle accuracy. In addition, objects with distinctive features, as well as partially occluded objects, are better detected because our method selects high value instances using classification uncertainty and intra-class diversity, further enhancing overall detection performance.

%% file: Table/Tab5.tex
\begin{table}[htbp]
  \centering
  \caption{Hyperparameter study. $\gamma$ is the hyperparameter used in inter-class diversity, $\beta$ is the hyperparameter used in localization and orientation uncertainty.}
  \resizebox{0.3\textwidth}{!}{
    \begin{tabular}{cc|cc}
    \toprule
    $\gamma$     & AP(\%) & $\beta$     & AP(\%) \\
    \midrule
    0.001 & -     & 0     & 29.31 \\
    0.005 & 31.96 & 0.25  & 33.97 \\
    0.01  & 35.79 & 0.5   & 35.79 \\
    0.05  & 33.15 & 0.75  & 31.80 \\
    0.1   & 22.62 & 1.0   & 33.93 \\
    \bottomrule
    \end{tabular}
  }
  \label{Tab_supp1}
  
\end{table}

%% file: Figure/img_add3.tex
\begin{figure}[t]
    \centering
    \begin{subfigure}{0.49\linewidth}
        \centering
        \includegraphics[width=\linewidth]{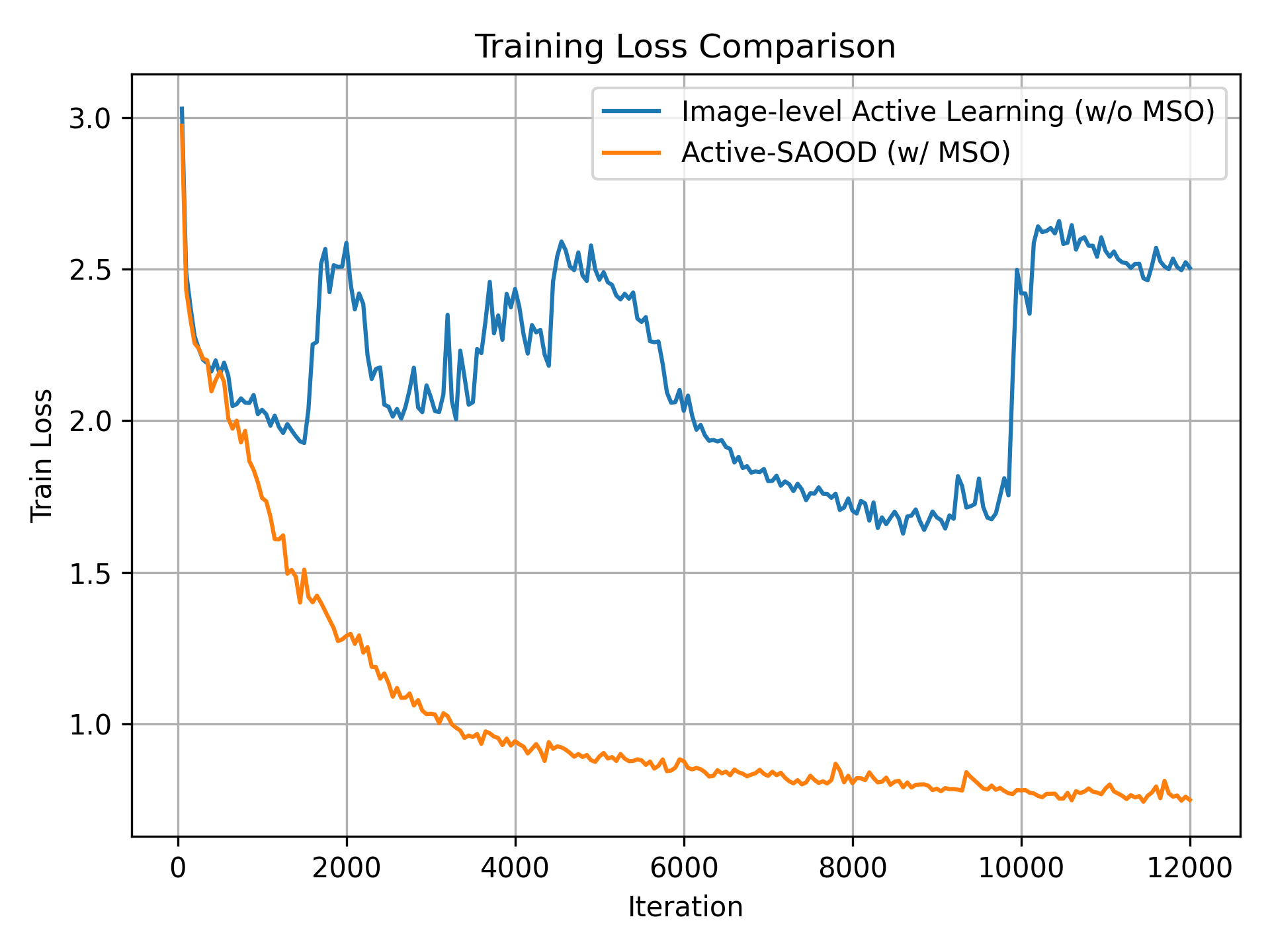}
        \caption{}
        \label{Fig_add3_a}
    \end{subfigure}
    \begin{subfigure}{0.49\linewidth}
        \centering
        \includegraphics[width=\linewidth]{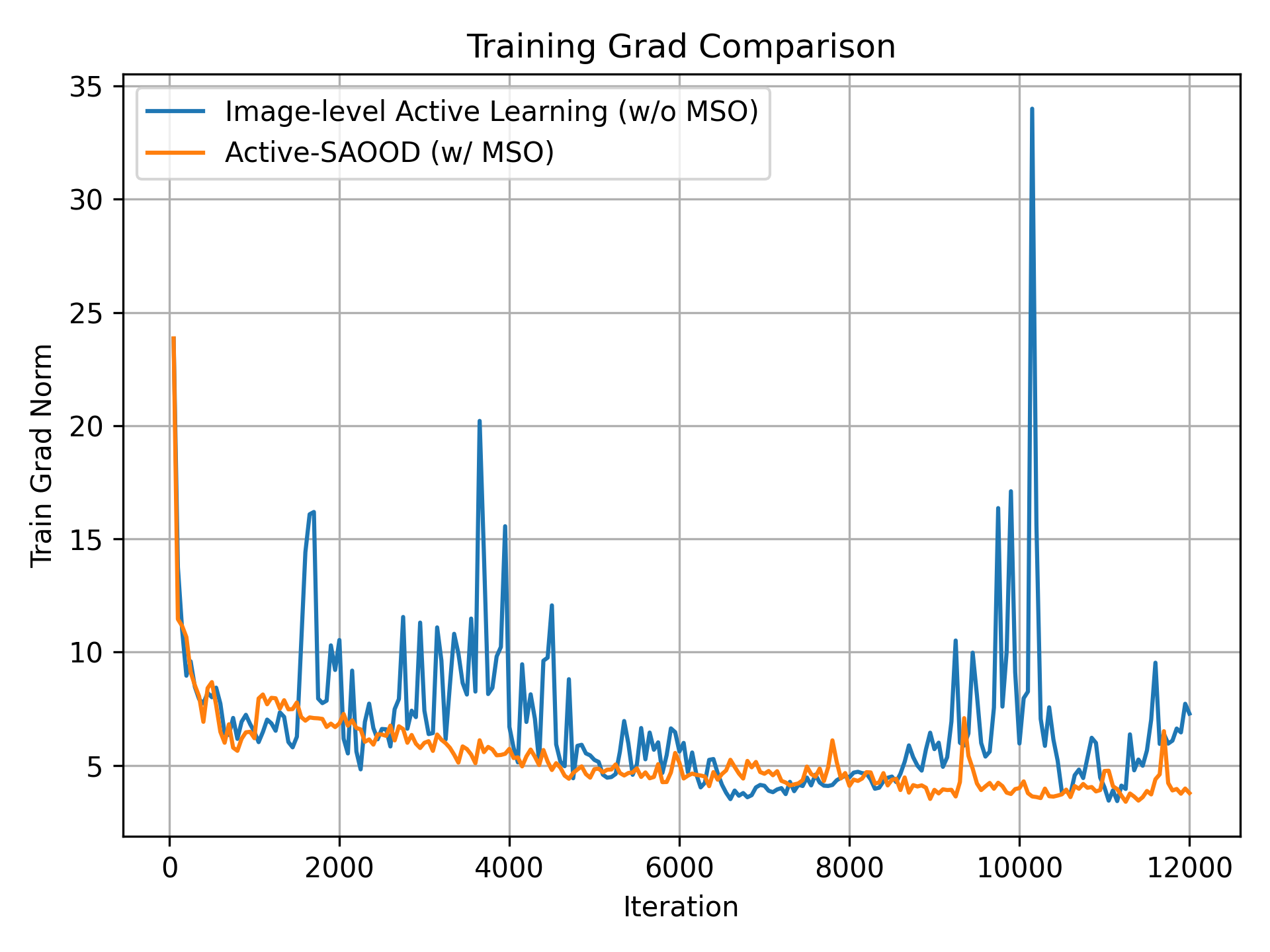}
        \caption{}
        \label{Fig_add3_a}
    \end{subfigure}
    \caption{Comparison with and without model state observation (MSO) strategy. (a) Training loss. (b) Training gradients. Traditional image-level active learning methods do not use MSO and directly select the most difficult instances, leading to unstable training. Active-SAOOD uses MSO to select appropriate instances according to the current model state, ensuring the training stability.}
    \label{Fig_add3}
    
\end{figure}

%% file: Figure/img_add1.tex
\begin{figure*}[t]
    \centering
    \begin{subfigure}{\textwidth}
        \centering
        \includegraphics[width=\textwidth]{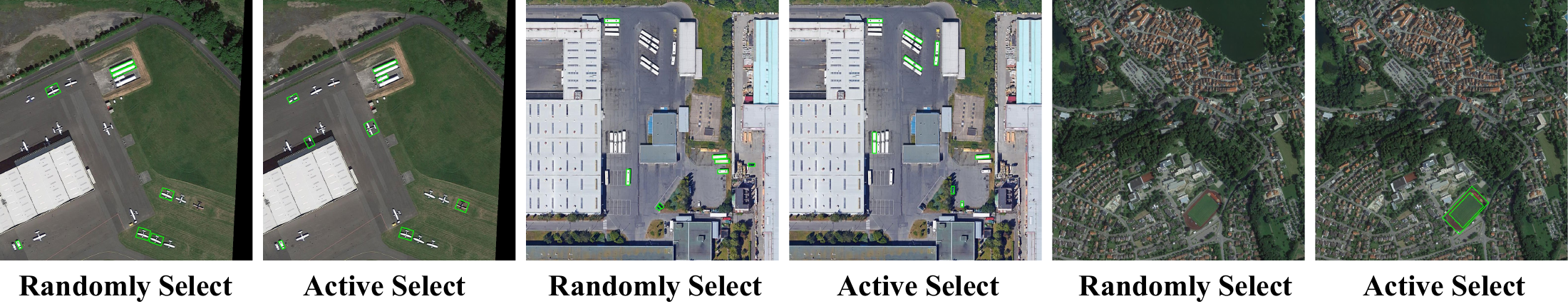}
        \caption{}
        \label{Fig_add1_a}
    \end{subfigure}

    \vspace{1em} 

    \begin{subfigure}{\textwidth}
        \centering
        \includegraphics[width=\textwidth]{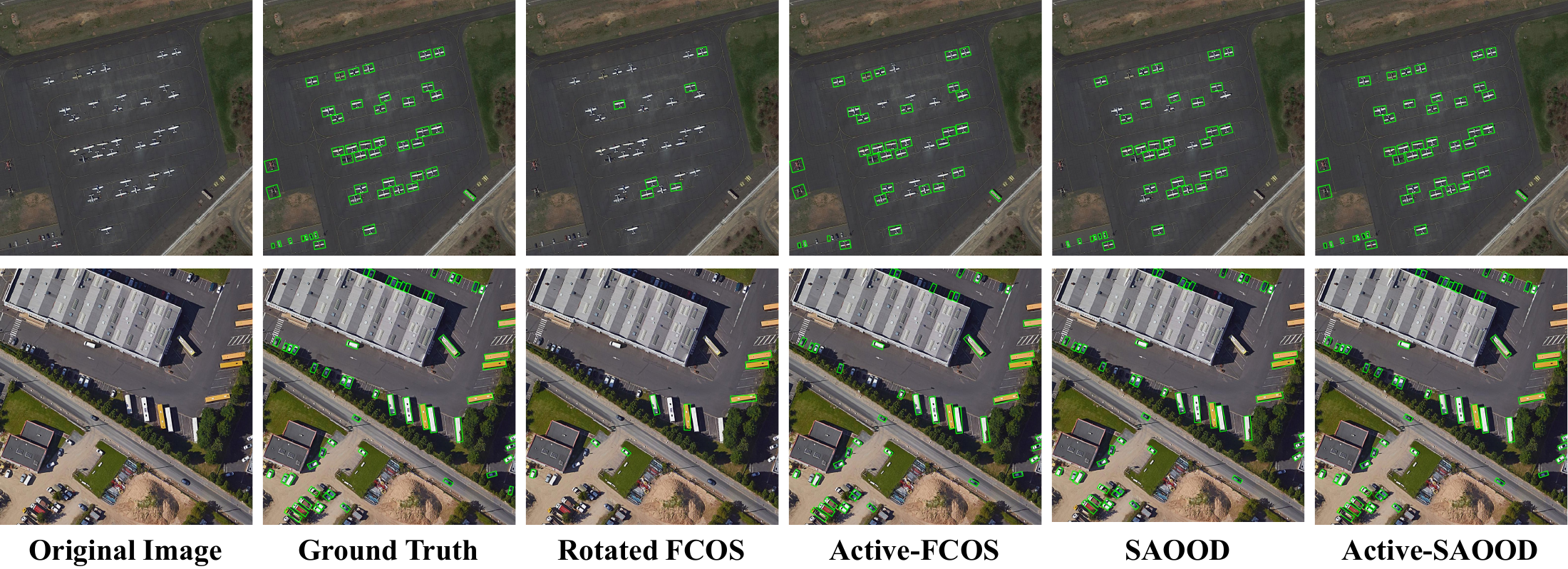}
        \caption{}
        \label{Fig_add1_b}
    \end{subfigure}

    \caption{Visualization analysis of instance selection and model performance. (a) Visualization comparing randomly annotated instances with actively selected instances. (b) Visualization comparing the detection results of models trained on randomly annotated instances versus those trained on actively selected instances.}
    \label{Fig_add1}
    
\end{figure*}

%% file: Figure/img_add2.tex
\begin{figure}[t]
    \centering
    \begin{subfigure}{0.5\textwidth}
        \centering
        \includegraphics[width=0.9\linewidth]{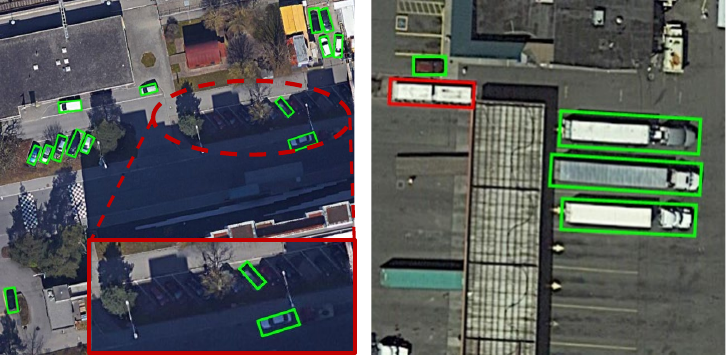}
        \caption{}
        \label{Fig_add2_a}
    \end{subfigure}

    \vspace{1em} 

    \begin{subfigure}{0.5\textwidth}
        \centering
        \includegraphics[width=0.9\linewidth]{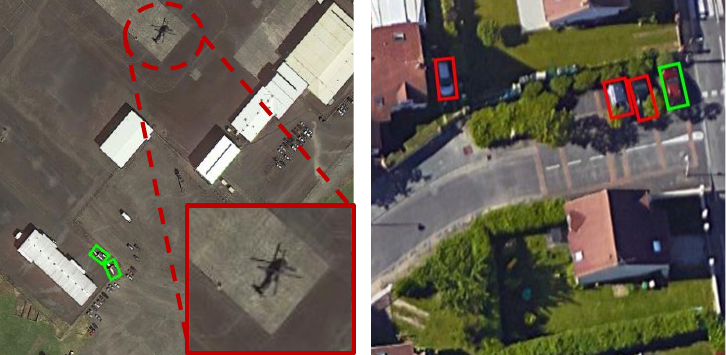}
        \caption{}
        \label{Fig_add2_b}
    \end{subfigure}

    \caption{Visualization analysis of the failure cases. Red boxes denote false positives, and green boxes denote true positives.}
    \label{Fig_add2}
    
\end{figure}